# Saliency Enhancement using Gradient Domain Edges Merging

Dominique Beaini, Sofiane Achiche, Alexandre Duperré, Maxime Raison

*Polytechnique Montreal, 2900 Edouard Montpetit Blvd, Montreal, H3T 1J4, Canada*

**Abstract**

In recent years, there has been a rapid progress in solving the binary problems in computer vision, such as edge detection which finds the boundaries of an image and salient object detection which finds the important object in an image. This progress happened thanks to the rise of deep-learning and convolutional neural networks (CNN) which allow to extract complex and abstract features. However, edge detection and saliency are still two different fields and do not interact together, although it is intuitive for a human to detect salient objects based on its boundaries. Those features are not well merged in a CNN because edges and surfaces do not intersect since one feature represents a region while the other represents boundaries between different regions. In the current work, the main objective is to develop a method to merge the edges with the saliency maps to improve the performance of the saliency. Hence, we developed the gradient-domain merging (GDM) which can be used to quickly combine the image-domain information of salient object detection with the gradient-domain information of the edge detection. This leads to our proposed saliency enhancement using edges (SEE) with an average improvement of the F-measure of at least 3.4 times higher on the DUT-OMRON dataset and 6.6 times higher on the ECSSD dataset, when compared to competing algorithm such as denseCRF and BGOF. The SEE algorithm is split into 2 parts, SEE-Pre for preprocessing and SEE-Post pour postprocessing.

**Nomenclature and Acronyms**

| | |
|---|---|
| $S$ | Saliency map normalized in the range [0, 1] |
| $M$ | Binary mask generated by a threshold on $S$ |
| $G$ | Binary ground truth of the saliency |
| $P$ | Precision of $S$ against $G$ |
| $P_{\max}$ | Maximum value of $P$ |
| $R$ | Recall of $S$ against $G$ |
| $F_m$ | F-measure on a given dataset |
| $\beta$ | Parameter for the F-measure set to 0.3 |
| $\overline{PR}$ | Average of the precision-recall curve |
| $!$ | Logical NOT operator |
| $AUC$ | Area under the true-false-positive curve |
| $MAE$ | Mean absolute error between $S$ and $G$ |
| $N$ | Total number of pixels in an image |
| $I_0$ | Input image normalized in the range [0, 1] |
| $I_R$ | The resulting image after solving the Laplacian $L_p$ |
| $\boldsymbol{E}$ | Gradient of the image $I_0$ |
| $L_p$ | Laplacian of the image $I_0$ |
| $\theta_E$ | Orientation of the gradient $\boldsymbol{E}$ |
| $K_{I \to E}$ | Complex kernel used to compute the gradient $\boldsymbol{E}$ from an image $I_0$ |
| $K_{E \to L}$ | Complex kernel used to compute the Laplacian $L_p$ from a gradient $\boldsymbol{E}$ |
| $E_{x,y}$ | $x$ and $y$ component of the gradient $\boldsymbol{E}$ |
| $C_0$ | Edges computed via an edge detection method |
| $\theta_C$ | Orientation of the edges |
| $E_p$ | Gradient modified using edges |
| $x, y$ | Horizontal and vertical axis |
| $M_E$ | Function used to merge the norm of the gradient with the edges |
| $M_\theta$ | Function used to merge the orientation of the gradient with the edges |
| $P_{I_{0 \to 1}}$ | Preparation step for the image |
| $P_{C_{0 \to 1}}$ | Preparation step for the edges |
| $S_C$ | Saliency map with enhanced contrast |
| $K$ | Contrast enhancement parameter |
| $\check{K}_{\nabla^2}$ | Zero-padded numerical Laplacian kernel |
| $\check{\delta}$ | Zero-padded numerical Dirac's delta |
| $\Re$ | Real part of the complex numbers |
| $\Im$ | Imaginary part of the complex numbers |
| $\mathcal{F}$ | Fourier transform |
| $\mathcal{F}^{-1}$ | Inverse Fourier transform |
| $\alpha$ | Exponent parameter to optimize |
| $\beta$ | Weight parameter to optimize |
| $\circ$ | Hadamard product (Element-wise multiplication) |
| $*$ | Convolution operator |
| $\cap$ | Intersection (And operator) |
| CNN | Convolutional neural network |
| FFT | Fast Fourier transform |
| GFC | Green function convolution |
| GDM | Gradient-domain merging |
| SEE | Saliency enhancement using edges |
| Post | For postprocessing |
| Pre | For preprocessing |
| CPU | Central processing unit |
| GPU | Graphics processing unit |

## 1. Introduction

Recent years, have seen great progress in solving binary problems in computer vision, such as edge detection [1–3] and saliency [1,4,5]. Saliency methods used different approaches based on multiple features, such as clustering and density [6–8], concavity [9], contrast filtering [10], and background detection [11], but did not use edges since there exist no methods of merging edges with saliency. Furthermore, since the arrival of convolutional neural network for saliency detection around the year 2015 [12], many highly effective algorithms were proposed such as MDF [12], DCL [5] and DSS [1,4]. The CNN were also used for edge detection such as RCF [2], with a work by Hou et al. in which they propose a unified framework to compute both saliency and edge detection [1], but without combining both results together.

Although there have been many research works on both saliency and edge detection, only a few propose to improve the saliency using edges [13–15]. Other works propose to use background detection [11], contrast enhancement and texture smoothing [16] to improve the results, but as benchmarked by Patel et Raman [13], those algorithms do not work well on the recent CNN models, since CNN models are more performant at detecting those features. In fact, only the denseCRF [15] and the boundary-guided BGOF [14] method are proven to improve the performance of CNN-based saliency detection [14]. This is supported by the simple fact that the saliency should be low outside the boundary and high inside it, which is exactly what denseCRF and BGOF work tries to optimize. However, they rely on an energy minimization of region segmentations instead of edge-based boundary conditions.

Hence, our objective is to develop a method that merges the results of edge detection algorithms with the results of saliency detection algorithms to improve the performance of the saliency maps. For this objective, we propose saliency enhancement using edges (SEE) which allows one to merge the results from top saliency methods with top edge detection methods.

Merging the region saliency maps with the edge maps is complex since they typically do not intersect, meaning that they cannot be combined by standard operations such as additions and multiplications. To solve this problem, our previous work showed that we can use the numerical 2D Green's function as convolutional kernels to extrapolate edges into surface information [17–20]. These convolutions allowed to quickly evaluate the probability of being inside a given contour [18] and to smooth an image according to the edges given by an edge detection technique [20]. The SEE approach proposed in the current paper assumes that the gradient of the saliency maps is similar in nature to the edges map. Hence, it uses the gradient-domain to merge those features, then it solves the gradient using a numerical Green's function convolution [20]. The SEE method is split into 2 parts: SEE-Pre to preprocess the image based on the edges and the SEE-Post to post-process the saliency map based on its edges. The optimal results occur when those 2 parts are used together.

## 2. Saliency enhancement using edges

For the saliency enhancement using edges (SEE) method, we propose to merge edges with the saliency and the image using gradient-domain merging (GDM). This SEE method is separated into 3 steps, the SEE-Pre which preprocesses the image, the SEE-Post which post-processes the image and the contrast enhancement.

The SEE-Pre works by blurring the non-salient regions and enhancing the contrast of the salient regions. Competitive approaches with the same goal were proven to work on standard saliency models [13–15], but not on the newest deep-learning algorithms [14]. The SEE-Post works by keeping only the salient regions that are bounded by salient edges, similarly to the most efficient available method BGOF [14].

The following sections will first explain the SEE method and will follow by detailing each step: the salient edge detection, the GDM, the SEE-Pre, the SEE-Post and the contrast enhancement.

### 2.1. The complete SEE method

An overview of the SEE method can be seen in Figure 1 where the edges and the salient edges $C_0$ and the image $I_0$ serve as the 2 inputs of the method. With the regular saliency being $S_I$, we observe that the SEE-Pre is first applied to get the enhanced saliency $S_{I_0R}$, followed by SEE-Post to obtain $S_{I_0RR}$. Finally the saliency map is normalized in the range $[0, 1]$ and the contrast is enhanced to get the saliency $S_{I_0RRC}$.

We can observe in Figure 1 that the precision/recall curve is almost perfect for the given example, with a precision $P$ near 100% at every point where the recall is $R$ is lower than 94%. This shows in the current example that the SEE method can significantly improve the resulting saliency map when both the preprocessing SEE-Pre and postprocessing SEE-Post methods are used together. More details on the precision/recall curve is given in section "2.5 Evaluation datasets and metrics".

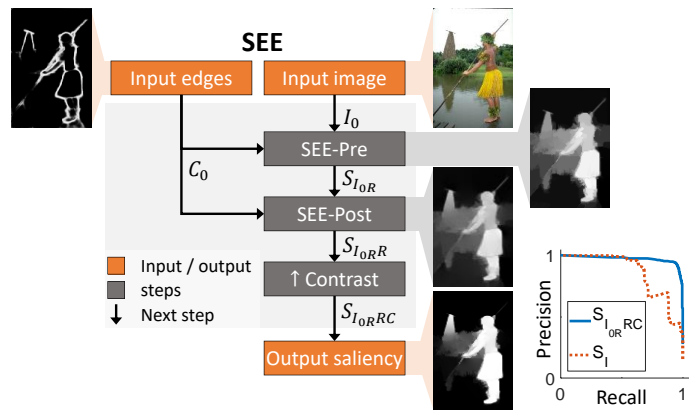

Figure 1. Diagram of the complete SEE method, which allows to improve the saliency via a combined image preprocessing and saliency map postprocessing. The images are examples of the results at each step based on the DRFI method with a precision/recall curve for the selected example where $S_I$ is the original saliency and $S_{I_0RRC}$ is the same saliency improved using our SEE method.

## 2.2. Salient edge detection

Before merging the edges in the gradient domain for our SEE approach, we must first understand which edges should be extracted from the image. Most edge detection methods focus on extracting all the edges from an image and its background [1,2,21] by basing their training on the BSDS500 dataset [22]. In the case of saliency improvement, this yields to undesired edges that are not useful for the saliency improvement.

For the task of salient object detection, the most used dataset for the training is MSRA10K with 10,000 images [23] or its earlier version MSRA-B with 5000 images [24]. Those saliency-based datasets include a ground-truth $G$ image with value 1 at every pixel inside the salient object and 0 at every pixel outside the salient object. By taking the boundaries of those $G$ images for the salient objects, we get a new ground-truth $G_{sal}$ for the salient edges. Using the new $G_{sal}$, we can retrain the edge detection models to detect only the salient edges.

We tried to use 2 different edge detection models to detect the salient edges by retraining or fine-tuning them with the $G_{sal}$. The first method, SE, uses random structured forest for learning [21,25] and it was one of the best method available before the introduction of the CNN [2,25]. The second method, RCF [2], uses deep CNN for learning [2], and it is one of the best edge detection method available [1,2]. Examples of results are shown in Figure 2. The training models on the BSDS500 were already available for both methods. For the MSRA10K, we retrained the SE method with 10k iterations and we fine-tuned the RCF method with 1k and 5k iterations. The SE method had only moderate success at eliminating non-salient edges, and with many missing salient edges. The finetuned RCF method with 1k iterations and 256 images per iteration performed a lot better by eliminating the non-salient edges and enhancing the salient edges. The same method with 5k iterations appears a lot cleaner, but also eliminates the salient edges which is undesirable.

A full test of the SEE method with different number of iterations of the RCF method suggest that 1k iterations of finetuning with 256 images per iterations are near optimal results for saliency improvement. Hence, we used this finetuning of RCF for the rest of the paper, which we will denote as RCF-MSRA10k-1k. Since the dataset has 10,000 images and that we used 8000 for training, each image was trained an average of 32 times, which is low enough to avoid an overfit.

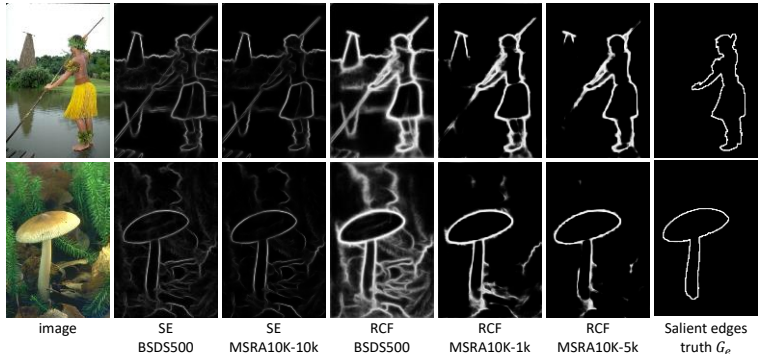

Figure 2. Comparison of SE and RCF edge detection methods trained or finetuned on the BSDS500 or MSRA10K datasets, with the number of iterations of retraining, using 2 example images from the ECSSD dataset. The salient edges will be used as inputs for the proposed SEE method.

## 2.3. Gradient-domain merging

Our previous work demonstrated that an image can be reconstructed from its gradient using a simple Green's function convolution (GFC) [18,20]. Since the convolution converges to the optimal solution even when a perturbation is added to the gradient, it showed that it can be used for robust edge-preserving blurring with any edge-detection method [20]. The blurring was done by gradient-domain merging (GDM) with the detected edges, followed by a Green's function convolution that computes the image associated to the modified gradient.

The current section presents the main equations that are used for gradient domain merging (GDM), which allows to merge the results of an edge detection with an image or saliency map. All the details for the equations can be found in previous work [17–20], with the most similarities to our previously proposed

Gradient and Laplacian solver [20]. We make use of complex numbers to avoid redundant equations and convolution kernel by representing the $x$ and $y$ axes in the same equation. However, in practice those equations can be separated into 2 different equations representing the real and imaginary parts.

Since the GDM requires its inputs to be in the gradient domain, the first step is to compute the gradient $\boldsymbol{E}$ of the image $I_0$ and its orientation $\theta$ using the complex right-derivative kernel $K_{I \to E}$ in equation (1), where $i$ is the imaginary number. It is possible to use 2 real kernels instead of $K_{I \to E}$, which would require 2 convolutions with the real/imaginary part of the kernel [18,20]. It is important to note that a zero-valued padding of 3 pixels must be added to $I_0$ to ensure continuity of the gradient and Laplacian at the border of the image [20].

$$K_{I \to E} = \begin{bmatrix} 0 & 0 & 0 \\ 0 & i-1 & 1 \\ 0 & -i & 0 \end{bmatrix}$$

$$\boldsymbol{E} = I_0 * K_{I \to E}$$

$$E_x = \Re(\boldsymbol{E}), \quad E_y = \Im(\boldsymbol{E})$$

$$|\boldsymbol{E}| = \sqrt{E_x^2 + E_y^2}, \quad \theta_E = \operatorname{atan}\left(\frac{E_y}{E_x}\right)$$

(1)

Using a merger function for the norm $M_E$ and another merger function for the orientation $M_\theta$, it is possible to merge the gradient $\boldsymbol{E}$ with the modified contour $C_1$, as seen in equation (2). This allows to obtain the modified gradient $\boldsymbol{E}_p$ with its orientation $\theta_p$ in the complex domain.

$$|\boldsymbol{E}_p| = M_E(C_0, \boldsymbol{E})$$

$$\theta_p = M_\theta(C_0, \boldsymbol{E})$$

$$\boldsymbol{E}_p = |\boldsymbol{E}_p| e^{i\theta_p}$$

(2)

Once the modified gradient $\boldsymbol{E}_p$ is computed, it is mandatory to go back from the gradient domain to the potential (or image) domain. Hence, we need to solve the gradient using the GFC method that we developed in previous work [20]. The GFC method is used since it is fast with around 2 ms of computation time for an image of size $800 \times 1200$ and it is optimally robust against perturbations or modifications [20].

Using the new modified gradient $\boldsymbol{E}_p$, we can compute the modified Laplacian $L_p$. The Laplacian is used since it is more straightforward to solve the Laplacian than the gradient using the GFC method [20]. The complex kernel $K_{E \to L}$ from equation (3) is used to compute the Laplacian $L_p$ in equation (4), where $\Re$ is the real part of the complex value.

$$K_{E \to L} = \begin{bmatrix} 0 & -i & 0 \\ -1 & i+1 & 0 \\ 0 & 0 & 0 \end{bmatrix}$$

(3)

$$L_p = \Re(\boldsymbol{E}_p * K_{E \to L})$$

(4)

The GFC method states that the Laplacian can be easily solved using the 2D Green's function. We first need to implement the matrices $\check{K}_{\nabla^2}$ and $\check{\delta}$ using equation (5), which are the zero-padded Laplacian and Dirac's kernels [20]. Then, we compute the optimal Green's function in the Fourier domain $\check{V}_{\text{mono}}^{\mathcal{F}}$ using equation (6), with $\mathcal{F}$ being the Fourier transform [20]. The value of $\check{V}_{\text{mono}}^{\mathcal{F}}$ can be pre-computed to reduce the computation time of the Laplacian solver.

$$\check{K}_{\nabla^2} \equiv \begin{bmatrix} 0 & -1 & 0 & \cdots & 0 \\ -1 & 4 & -1 & & \\ 0 & -1 & 0 & & \\ \vdots & & & \ddots & \\ 0 & & & & 0 \end{bmatrix}_{\underbrace{\hspace{2cm}}_{size(I)}} \quad \check{\delta} \equiv \begin{bmatrix} 0 & 0 & 0 & \cdots & 0 \\ 0 & 1 & 0 & & \\ 0 & 0 & 0 & & \\ \vdots & & & \ddots & \\ 0 & & & & 0 \end{bmatrix}_{\underbrace{\hspace{2cm}}_{size(I)}}$$

(5)

$$\check{V}_{mono}^{\mathcal{F}} = \frac{\mathcal{F}(\check{\delta})}{\mathcal{F}(\check{K}_{\nabla^2})} \qquad (6)$$

Finally, the Laplacian can be solved with a convolution in the Fourier domain [20,26] using equation (7), where $\mathcal{F}$ and $\mathcal{F}^{-1}$ are the Fourier transform and its inverse, $\mathcal{R}$ is the real part of a complex number and ∘ is the element-wise product. The resulting $I_R$ has a constant value of $-c$ on all of its borders, which will be set to zero by defining adding the integration constant $c$ to $I_R$ [20]. Also, the resulting $I_R$ is cropped to match the initial size of $I_0$ before the padding.

$$I_R = \mathcal{R}\left(\mathcal{F}^{-1}\big(\mathcal{F}(L_p) \circ \check{V}_{mono}^{\mathcal{F}}\big)\right) + c \qquad (7)$$

A diagram representing the previous steps is presented in Figure 3, where $C_0$ is the salient edges and $I_0$ is either the saliency map or the RGB image. In the current work, $C_0$ is computed via the method RCF-MSRA10k-1k. The gradient step is computed with (1), the Laplacian with (4) and the GFC with (7). The preparation steps are flexible and the merging functions $M_E$ and $M_\theta$ (2) vary according to the desired application and they will be explained in later sections.

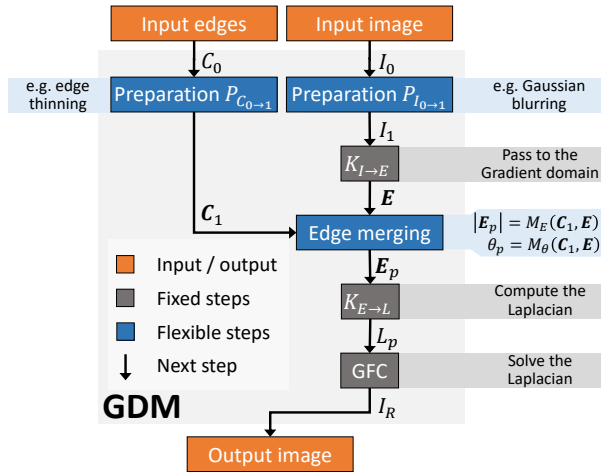

Figure 3. Diagram of the core GDM function, allowing to merge image/saliency information with edges information by passing to the gradient domain and solving the Laplacian equation to go back to the image domain.

In summary, this section explained how to compute the GDM by solving the modified gradient $E_p$ using the Laplacian $L_p$ (4), the pre-computed Fourier-domain Green's function (6), and the Fourier-domain convolution (7). This allows to merge the image or saliency information with the salient edges by using the gradient domain, as shown in Figure 3. As stated in our previous work, this computation is fast and easy to implement [18,20], since most computer vision libraries such ®MATLAB [27] and OpenCV [28] implement fast Fourier transform (FFT) algorithms. Also, the FFT algorithms are also available on the graphical processing unit (GPU) which allows for faster computation. The estimated computation time for a single channel 400x400 image is 1.5ms when eliminating overhead on the GPU ®Nvidia GTX 1080-Ti.

*2.4. The SEE method*

The GDM method explained in the previous sub-section can be used at different stages for the improvement of this saliency, with some variations outlined by the "flexible steps" in Figure 3. This section explains how GDM can be used for saliency enhancement using edges (SEE). First, we show how GDM is used for postprocessing of the saliency map (SEE-Post) and preprocessing of the image (SEE-Pre). Then, we show how to combine them into the general SEE method.

*2.4.1. The SEE-Post method*
The first part of the SEE method consists of the postprocessing of the saliency map SEE-Post, which tends to enhance its intensity inside the salient edges and to reduces it outside them.

Since the SEE-Post acts as a postprocessing of the saliency result, then the input $I_0$ of the GDM is replaced by the saliency $S_I$ computed on the image $I$. The input $C_0$ of the GDM is the salient edge computed using RCF-MSRA10k-1k. The output of GDM is the reconstructed saliency $S_R$. Finally, the SEE-Post method averages $S_I$ and $S_R$, resulting in the final output $S_{IR}$. The output $S_{IR}$ combines the higher precision of $S_R$ with the higher recall of $S_I$ for a better overall result. The entire procedure is explained in the Figure 4, where is the example saliency $S_I$ is given by the DRFI method [24]. The final results are given later in the section 3 and the precision-recall curves are in the section 4.

Although our SEE-Post uses different saliency methods, DRFI is chosen for the examples since it allows to better visualize the improvement of the saliency map. On Figure 4, we observe how the intensity of the background is significantly reduced while the intensity of the salient person is enhanced. The improvement is shown in the precision/recall curve of Figure 4 where the $S_{IR}$ curve has better or equal precision than $S_I$ at every point and $S_{IR}$ is shown have an almost steady precision for any recall value.

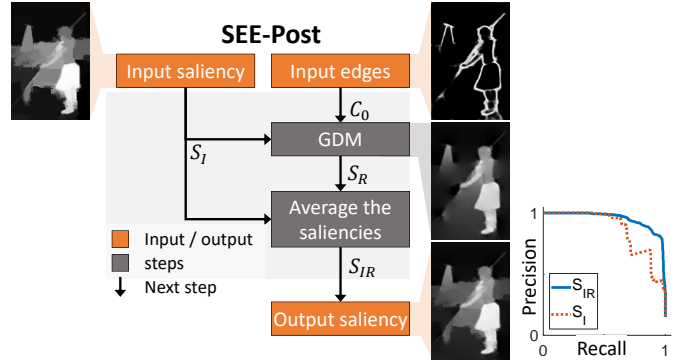

Figure 4. Diagram of the SEE-Post method, which allows to postprocess a saliency map based on the salient edge detection. The images are examples of the results at each step based on the DRFI method with a precision/recall curve for the selected example.

Now we need to understand the preparation steps for the edges $P_{C_0 \to 1}$ and for the saliency $P_{S_0 \to 1}$ that are used in order to obtain those results. For optimal results, we want the edges to be as thin as possible to avoid blurring the edges of the saliency map and to have a more accurate tracing. Hence, the $P_{C_0 \to 1}$ applies the non-maximal suppression (NMS) algorithm provided with the SE edge detection [21,25] on their ®GitHub [29]. However, edges too thin will not necessarily intersect with the gradient, which means that $P_{C_0 \to 1}$ adds a dilation after the thinning using the *imdilate* function in MATLAB® with a disk of 3-pixels diameter.

Furthermore, the $P_{S_0 \to 1}$ adds a blurring to the saliency map to ensure the smoothness of the reconstruction, as proposed by our previous work for the painting effect [20]. We use a Gaussian blur with standard deviation $\sigma = 3$ pixels to compute $S_1$, which helps ensure a smooth intersection between the gradient of $S_1$ and the thinned edges $C_1$.

The next step is to define the functions $M_E^{Post}$ and $M_\theta^{Post}$ from equation (2) to determine how the gradient merging is done. First, we define the merger function for the norm $M_E^{Post}$ according to equation (8), which is inspired by the edge contrast enhancement and painting effect from our previous work [20]. The element-wise product "∘" allows to eliminate every gradient that is not nearby a thin edge, while the square-root allows to preserve $|E|$ in case $|E| = C_1$.

$$\left|E_p^{Post}\right| = M_E^{Post}(C_1, E) = \sqrt{C_1 \circ |E|} \tag{8}$$

Then we define the merger function for the orientation $M_\theta^{Post}$ according to equation (9), which always returns the value of $\theta_C$, but shifts it by $\pi$ if the projection of $E$ on $C_1$ returns a negative value. $\theta_C$ is defined as the orientation perpendicular to the thin edges. This is based on the fact that our previous work [18] proved that the orientation of dipoles must be perpendicular to the contour $C_1$ to give the probability that any pixel is inside the given $C_1$, which is closely related to the saliency. Furthermore, the shift by a value of $\pi$ is due to the 2 different possible orientation of dipoles, which must ideally be optimized [18]. Since the optimization is computationally heavy, choosing the orientation in the same direction as $E$ yields to satisfactory results.

$$\theta_p^{\text{Post}} = M_\theta^{\text{Post}}(C_1, \boldsymbol{E}) = \begin{cases} \theta_C & \cos(\theta_C - \theta_E) \geq 0 \\ \theta_C + \pi & \cos(\theta_C - \theta_E) < 0 \end{cases} \quad (9)$$

In summary, the general steps for SEE-Post method are presented in Figure 4, with the specific flexible GDM steps presented in the list below.
- Contour preparation step $P_{C_{0\to 1}}$
  - Control thinning using NMS, followed by dilation with a disk of 3-pixel diameter.
- Saliency preparation step $P_{S_{0\to 1}}$
  - Smoothing using a normalized kernel Gaussian kernel with $\sigma = 3$
- Norm merging function $M_E^{\text{Post}}$ defined in equation (8).
- Orientation merging function $M_\theta^{\text{Post}}$ defined in equation (9).

*2.4.2. The SEE-Pre method*

In addition to the postprocessing method, the proposed SEE approach also offers a preprocessing method SEE-Pre for the improvement of the computed saliency map. It works by using salient edges to generate a new image where most of the background is eliminated, which helps the saliency method generate more accurate saliency maps.

Since the SEE-Pre acts as a preprocessing of the input image, then the input $I_0$ of the GDM is the original image $I_0$ and the input $C_0$ of the GDM is the salient edge computed using RCF-MSRA10k-1k. The output of GDM is the reconstructed image $I_R$. Then the SEE-Post method runs twice the saliency algorithm to obtain $S_{I_0}$ and $S_{I_R}$ and averages both outputs, resulting in the final saliency $S_{I_0R}$. The averaging allows to merge the better object recall of $S_{I_R}$ with the better boundary precision of $S_{I_0}$. The entire procedure is explained in the Figure 5, where the example saliency maps $S_{I_0}$ and $S_{I_R}$ images are given by the DRFI method [24].

Once again, the SEE-Post works with different saliency methods, but DRFI is chosen to better observe the improvements generated by the SEE method. On Figure 5, we observe how the intensity of the person is significantly enhanced compared to the background. The improvement is shown in the precision/recall curve of Figure 5 where the $S_{I_0R}$ curve has better or equal precision than $S_I$ at every point. The curve is steady and somewhat similar to the one from SEE-Post in Figure 4, although the contrast between the salient person and the background is higher with the SEE-Pre than with the SEE-Post.

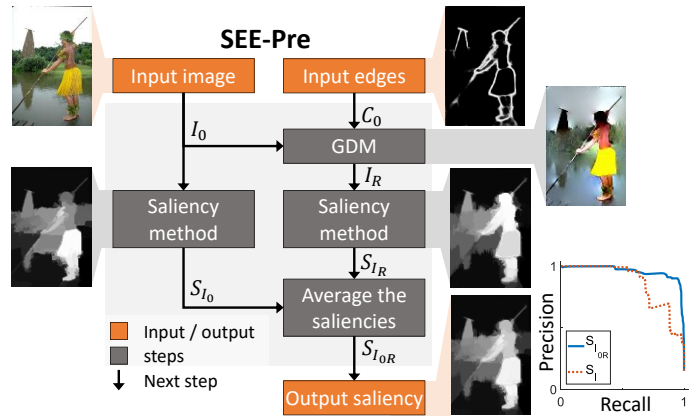

Figure 5. Diagram of the SEE-Pre method, which allows to preprocess an image based on the salient edge detection for a more accurate saliency map. The images are examples of the results at each step based on the DRFI method with a precision/recall curve for the selected example.

For the GDM step of Figure 5, we can first ignore the preparation steps for the input edges $P_{C_{0\to 1}}$ and for the input edges $P_{I_{0\to 1}}$, meaning that $C_1^{\text{Pre}} = C_0^{\text{Pre}}$ and $I_1^{\text{Pre}} = I_0^{\text{Pre}}$.

Now we need to define the functions $M_E^{\text{Pre}}$ and $M_\theta^{\text{Pre}}$ from equation (2) to determine how the gradient merging is done. The merger function for the norm $M_E^{\text{Pre}}$ is defined in equation (10), which is similar to equation (8), but averaged with the field $|\boldsymbol{E}|$ to preserve part of the texture and ensure the closure of the regions.

$$|\mathbf{E}_p^{\text{Pre}}| = M_E^{\text{Pre}}(C_1, \mathbf{E}) = \frac{\sqrt{|C_1| \circ |\mathbf{E}|} + |\mathbf{E}|}{2} \tag{10}$$

Then we define the merger function for the orientation $M_\theta^{\text{Pre}}$ simply uses the same angle as the gradient in order to preserve the texture and coloring of the image, as defined in equation (11).

$$\theta_p^{\text{Pre}} = M_\theta^{\text{Pre}}(C_1, \mathbf{E}) = \theta_E \tag{11}$$

In summary, the general steps for SEE-Pre method are presented in Figure 5, with the specific flexible GDM steps presented in the list below.
- Contour preparation $P_{C_{0\to1}}$ and image preparation $P_{I_{0\to1}}$ steps are ignored.
- Norm merging function $M_E^{\text{Pre}}$ defined in equation (10).
- Orientation merging function $M_\theta^{\text{Pre}}$ returns simply $\theta_E$ according to (11).

*2.4.3. Contrast enhancement*

Due to the use of inclusion probabilities and multiple averaging, it is preferable to enhance the contrast of the resulting image as suggested by our previous work on improving the probability of inclusion [18]. Hence, we compute the contrast enhanced saliency $S_C$ by using the smooth-step provided in equation (12) and based on Hermite polynomial [18,30], where $S$ is any saliency map and $S_C$ is the contrast enhanced saliency map. This polynomial enhances the value of any $S > 0.5$ and reduce the value of any $S < 0.5$. For the example of Figure 1, we use $K = 4$, with more detailed explanations to follow.

$$S_C = S^{K+1} \sum_{k=0}^{K} \binom{K+k}{k}\binom{2K+1}{K-k}(-S)^k \tag{12}$$

In the smooth-step equation (12), $K$ is a parameter representing the intensity of the contrast enhancement, with $K = 0$ representing no contrast enhancement and $K \to \infty$ representing a step function that sets to 0 any value of $S < 0.5$ and to 1 any value of $S > 0.5$. In our current work, we use $K = 4$ since it helps improve the mean absolute error defined in equation (20), without any noticeable impact on the other parameters Examples of the polynomial (12) for $K = \{2, 4\}$ are presented in equation (13). Furthermore, Figure 6 allows to better visualize the effect of the smooth-step function.

$$\begin{aligned} S_{C_{(K=2)}} &= 6\,S^5 - 15\,S^4 + 10\,S^3 \\ S_{C_{(K=4)}} &= 70\,S^9 - 315\,S^8 + 540\,S^7 - 420\,S^6 + 126\,S^5 \end{aligned} \tag{13}$$

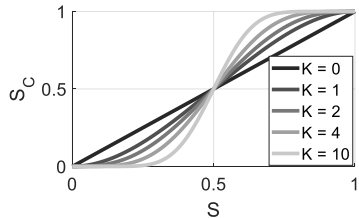

Figure 6. Smooth-step function (12) with different values of $K$.

In the Figure 1 representing the whole SEE method, the saliency $S_{I_0RRC}$ is the application of the equation (13) on the saliency $S_{I_0R}$ resulting from the combination of SEE-Pre and SEE-Post.

*2.5. Evaluation datasets and metrics*

To properly evaluate our proposed SEE algorithm, we need to use standard datasets and metrics. For the datasets, we use the MSRA10K [23] for training purposes, which is an extension of the previous MSRA-B [24] dataset. The MSRA10K is used for training since it has the largest number of images (10,000). It is also one of the easiest in terms of performance which makes it is harder do discriminate between different algorithms, and it is the most used for training purposes [4].

For evaluation purposes, we use the following 3 datasets: ECSSD with 1000 images [31,32], PASCAL-S [33] with 850 more complex images and DUT-OMRON with the most complex 5168 images [34]. These datasets are used since they are among the standard in the literature for test evaluation purposes and they are used for the BGOF method [14].

For the comparison with other techniques, the parameters that are evaluated are the precision $P$, the recall or true positives $R$ and the false positives $^!R$ [6,35]. Those parameters are evaluated for 256 levels of thresholds on the saliency map $S$, which allows to plot the precision-recall $PR$ curve. At each threshold level, a binary mask $M$ is generated and compared to the binary ground-truth $G$. From the $PR$ curve, one can evaluate the average $\overline{PR}$, the F-measure $F_m$ and the maximum precision $P_{\max}$. All those parameters are defined in equations (14)-(18), where $\beta = 0.3$ is a constant that allows to add more weight to the precision, $^!$ is the logical NOT operator, $\cap$ is the logical AND operator and $\sum$ is the sum over every pixel [6,35].

$$P = \frac{\sum M \cap G}{\sum M} \quad (14)$$

$$R = \frac{\sum M \cap G}{\sum G}, \quad ^!R = \frac{\sum M \cap {^!G}}{\sum {^!G}} \quad (15)$$

$$P_{max} = max(P) \quad (16)$$

$$F_m = max\left(\frac{(1+\beta^2)(P\,R)}{\beta^2\,P\,R}\right) \quad (17)$$

$$\overline{PR} = \int P\,dR \quad (18)$$

Other important information is the area under the curve (AUC) of the true-false-positive curve, and the mean absolute error (MAE) given respectively in equations (19) and (20), where $S$ is the saliency map normalized to $[0, 1]$ and $N$ the total number of pixels.

$$AUC = \int R\,d^!R \quad (19)$$

$$MAE = \frac{1}{N}\sum |S - G| \quad (20)$$

From all those parameters, the most used in the literature are the precision-recall $PR$ curve, the F-measure $F_m$ and the mean absolute error $MAE$. Hence, those parameters will be used to be compared with other methods from the literature. Additionally, we use the maximum precision $P_{\max}$, the mean precision-recall $\overline{PR}$ and the area under curve AUC to show that our approach improves many different measures.

**3. Results**

The complete SEE method was presented within the last section, but with only a single image example for the results. This section will show how the SEE methods on different images of the datasets ECSSD [31,32], PASCAL-S [33] and DUT-OMRON [34]. Different image examples will be provided, along with a benchmarking of the computation time.

*3.1. Results on different images*

The saliency methods on which the SEE is tested are DSS [1,4], DCL+ [5] (which combines DCL with denseCRF [15]), DRFI [24], RBD [11] and MC [36]. As we observe on the different examples in Figure 7, the SEE method significantly reduces the background values while enhancing the foreground values. Also, it partially fills some missing regions of the salient objects. Since a good saliency map is one where all the foreground values are higher than the background values, then we understand visually how our SEE method improves the best tested algorithms such as DCL [5] and DSS [4].

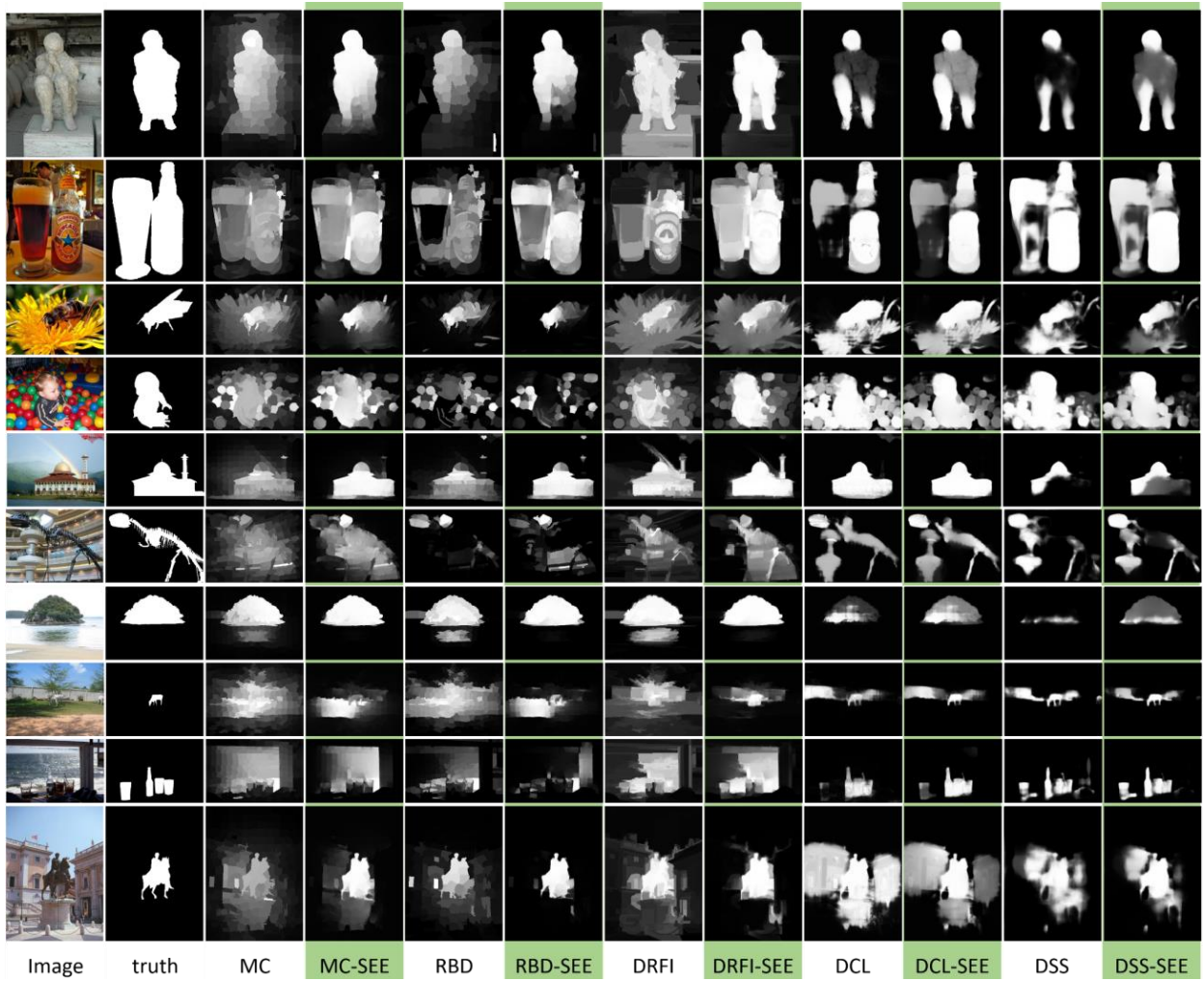

Figure 7. Comparison of the results from 5 different SoA methods (MC, RBD, DRFI, DCL and DSS) and their improvement using our SEE algorithm highlighted in green. The examples are chosen as some of the most difficult images in the datasets.

## 3.2. Computation time

Since our proposed SEE approach combines a SoA salient object detection with an SoA edge detection, then it is necessarily slower than the salient object detection alone. However, since many methods use denseCRF to improve their results [4,5], our method might perform faster by removing this layer. For our benchmarks, we use a graphic processing unit (GPU) Nvidia® GTX 1080 Ti and a central processing unit (CPU) Intel® i7-6700K.

For the GDM part, which is fundamental to the SEE algorithm, the computation time is around 1.5ms per RGB channel of size $400 \times 400$ on the GPU [20], which means that all the required GDM take around 6ms per image. This is negligible in the full computation time.

For the edge detection time, we use the RCF method, which takes around 100ms to compute a multiscale result [2]. For the salient object detection, the DCL method takes around 1000ms (due to the segmentation), while the DSS method takes around 80ms [4]. Both methods require an additional 400ms for using denseCRF to improve their saliency map [4]. Our method allows to remove the denseCRF but needs to compute the saliency twice. Hence, there is a computation time improvement for DSS, but not for DCL. The total computation time of the DSS-SEE method is less than 300ms while it is around 500ms for DSS-denseCRF. The total computation time of the DCL-SEE method is around 2000ms while it is around 1400ms for DCL-denseCRF.

## 4. Literature comparison and discussion

This section will perform a thorough benchmarking of the SEE-Post and the SEE methods on the datasets ECSSD [31,32], PASCAL-S [33] and DUT-OMRON [34]. The benchmarking includes measurements for the improvements over the saliency methods DSS [1,4], DCL+ [5],which combines DCL with denseCRF [15], MDF [12], DRFI [24], RBD [11] and DSR [8]. It also includes a comparison to state-of-the-art (SoA) methods for saliency maps improvement SO [11], denseCRF [15] and the most performant BGOF [14].

*4.1. Improvement of the saliency maps*

By using 3 different datasets and 7 SoA saliency methods, the current section shows that our proposed SEE approach allows to significantly improve the saliency results of many SoA algorithms, including the most recent CNN-based methods such as MDF, DSS and DCL+.

On Figure 8, we can observe that both the SEE-Post and the SEE methods allow to improve all 5 metrics defined in section 2.5 on the ECSSD (E), PASCAL-S (P-S) and DUT-OMRON (D-O) datasets. In fact, the only metrics where SEE reduces the performance is the $P_{\max}$ for the DSS method.

The improvement is high enough that some less performant methods can outperform methods that are significantly better. Hence, we observe that MC with SEE outperforms DRFI on many measures and that DRFI with SEE outperforms MDF on many measures.

Also, we observe on Figure 8 that some methods receive a higher boost of performance than similar performing methods, since they are naturally more adapted to the SEE method. For example, DSR is less improved than similar performing methods such as RBD, since the gradient of the RBD method merges better with the salient edge detection. Also, we note that the best regular method is DSS, but the best method is DCL+ when using the SEE algorithm. Again, this is because DCL+ merges better in the gradient domain, which gives it a bigger boost, especially for $F_m$, $\overline{PR}$ and $P_{\max}$.

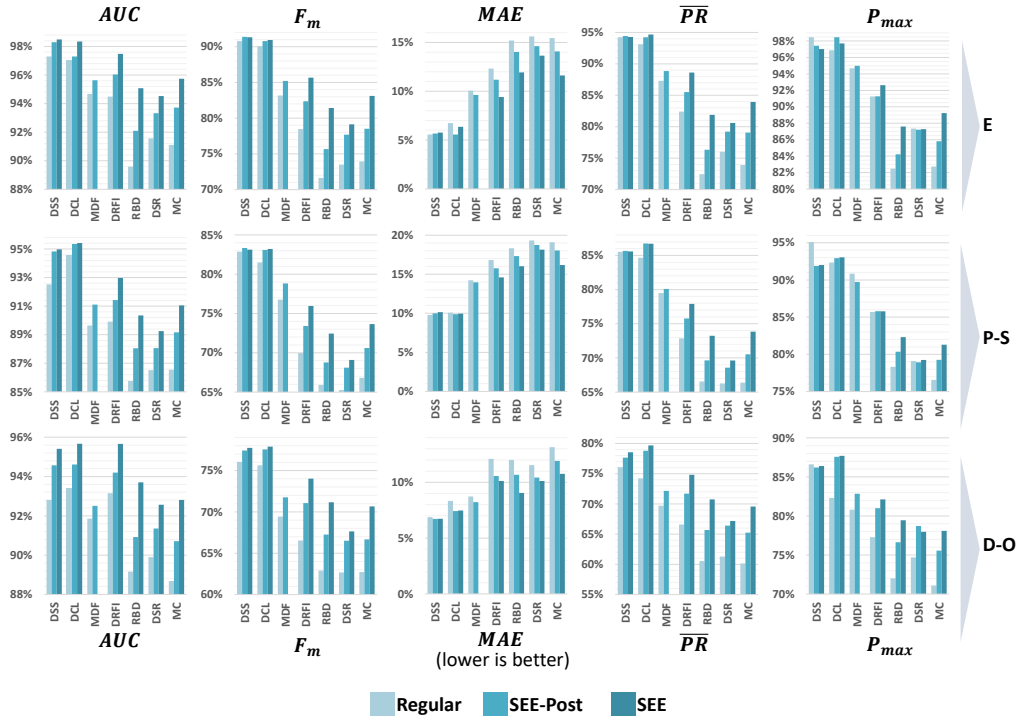

Figure 8. Comparison between the regular saliency methods, the enhanced saliency using our SEE-Post algorithm and the enhanced saliency using our SEE algorithm. A higher percentage is better for $AUC$, $F_m$, $\overline{PR}$ and $P_{\max}$, but a lower percentage is better for $MAE$. The 3 datasets used are ECSSD (E), PASCAL-S (P-S) and DUT-OMRON (D-O).

13For the same methods, we can observe the precision-recall $PR$ curves on Figure 9 on the same 3 datasets. We observe that for any non-CNN based saliency method, the improvement of the SEE method is very high at every point of the $PR$ curves. However, for the CNN-based methods, the regular method sometimes outperforms the SEE improvement at low recall ($R < 0.5$), but never at high recall ($R > 0.5$). Also, the $PR$ curve is a lot flatter with the SEE method, meaning that the precision is almost constant for every recall that is not too high ($R < 0.8$). This is reflected in the improvement of the $AUC$ and the $F_m$ parameters in Figure 8. The flat curve means that the saliency is more robust with the SEE method and will lead to easier thresholding and more robust thresholding. Furthermore, we observe again that the DCL with SEE outperforms the DSS with SEE, although DSS outperforms DCL.

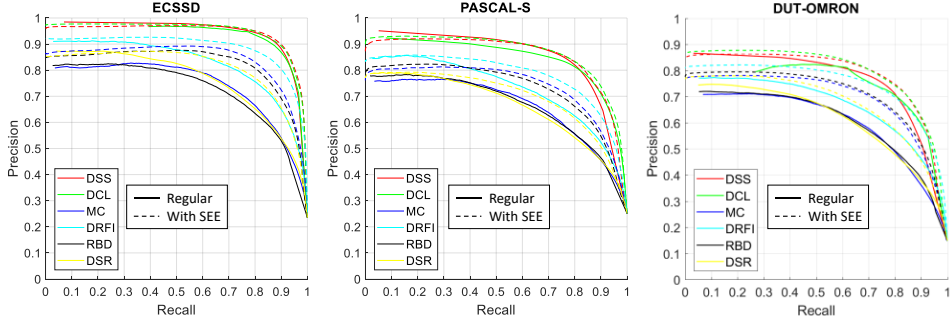

Figure 9. Precision-recall curves comparing 6 different methods (DSR, RBD, DRFI, MC, DCL and DSS) with and without our SEE method on 3 different datasets.

*4.2. Comparison to saliency improvement methods*

The previous section shows that our SEE method works well for its purpose of improving the saliency, and this section pursues by showing that it performs far better than any other algorithm with the same goal.

In fact, the improvement of SEE over $F_m$ is on average 6.6 times better than the nearest competing algorithm BGOF [14] on ECSSD and 3.4 times on DUT-OMRON, as observed on Figure 10. Also, other methods such as SO [11] and denseCRF [15] are even further behind. Since $F_m$ is the most universally used measure [4–6,24,35,37], this is an important achievement for the SEE method. Furthermore, the performance is better for $MAE$, another widely used indicator [5,6,24]. However, MAE is not as important as $F_m$ since it can easily be improved by enhancing the contrast of the saliency map, while $F_m$ is non-trivial to improve and indicates a direct improvement in the PR curve.

In summary, we observe that the proposed SEE method, and even the proposed SEE-Post are a vast improvement compared to any other algorithm present in the literature. Hence for computation time reasons, one can chose to not use the SEE-Pre method, especially for the high performance methods such as DCL and DSS where most of the improvements happens with the SEE-Post part of our approach.

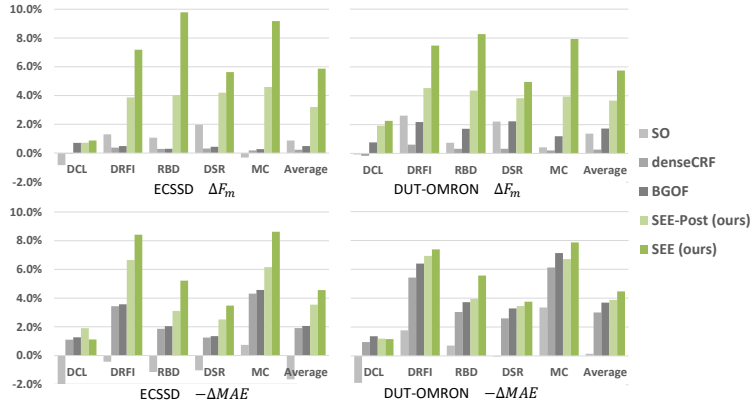

Figure 10. Comparison of the improvement over $F_m$ and $MAE$ for different SoA saliency improvement methods.

## 5. Conclusion

The objective of this paper was to develop a novel method of merging the edges with the saliency maps to improve the performance of the salient object detection. It is the first work that allows to combine the best advances in edge detection with the best advances in salient object detection. As seen in Figure 7, it works intuitively by reducing the values of the saliency map outside salient edges and enhancing it inside them, which is similar to how a human will perceive a salient object by its inclusion within boundaries. When compared to other methods of improving saliency maps with Figure 10, the SEE method shows an average improvement of the F-measure $F_m$ 3.4 times more than the BGOF on the DUT-OMRON dataset and 6.6 times on the ECSSD dataset, and an improvement of the mean absolute error significantly better than all its competitors. We also showed how the SEE method improves by a high margin the precision-recall curve and some other measures such as the $AUC$, $\overline{PR}$ and $P_{\max}$.

We believe that the proposed SEE method will have an important impact for the binary problems of computer vision since SEE is the first method that allows to merge edge detection methods with saliency detection methods for improved results. A limitation of the method is that it needs 2 different neural networks trained separately which requires to optimize the parameters of both networks simultaneously and which increases the computation time. Future work can focus on integrating the SEE method directly inside a neural network so that a single network is used, and all the parameters are optimized during the training process.

## Acknowledgment

The authors are grateful to NSERC, through the discovery grant program RGPIN-2014-06289, and FRQNT/INTER for their financial support.